\documentclass{article}
\usepackage{amsmath}
\usepackage{multirow}

\usepackage[preprint]{neurips_2026}

\usepackage[utf8]{inputenc} 
\usepackage[T1]{fontenc}    
\usepackage{hyperref}       
\usepackage{url}            
\usepackage{booktabs}       
\usepackage{amsfonts}       
\usepackage{nicefrac}       
\usepackage{microtype}      
\usepackage{xcolor}         
\usepackage{graphicx}
\usepackage{array}
\usepackage{bbm}
\usepackage[inline]{enumitem}
\title{Open-Vocabulary Domain Unlearning}

\author{%
  Sumanth Udupa \\
  The University of Queensland \\
  \And
  Mehrtash Harandi \\
  Monash University \\
  \AND
  Yadan Luo \\
  The University of Queensland \\
  \AND
  Mahsa Baktashmotlagh \\
  The University of Queensland \\
}

\definecolor{my_purple}{HTML}{9903F0}

\begin{document}

\maketitle

\begin{abstract}
Vision-Language Models (VLMs) exhibit remarkable zero-shot generalization, yet they often encode unwanted or hazardous stylistic domains such as idealized textbook diagrams in medical AI or cartoon vehicles in autonomous driving. Approximate Domain Unlearning (ADU) aims to selectively erase a model's recognition of a target visual domain while preserving accuracy on the remaining domains. However, existing ADU methods operate under a flawed closed-vocabulary assumption: they evaluate unlearning solely on the specific object classes seen during the unlearning fine-tuning phase. Consequently, these methods do not unlearn the domain itself; they merely overfit to seen class-domain pairs, leaving the domain easily recognizable for unseen classes and providing a false sense of removal. We argue that true domain erasure must be class-agnostic. To address this, we formalize \textbf{Open-Vocabulary Domain Unlearning (OVDU)}, a rigorous protocol that mandates domain forgetting must transfer to held-out classes. To solve the OVDU challenge, we propose a surgical parameter-editing framework. First, a Fisher Information mask isolates domain-sensitive weights, mathematically protecting foundational zero-shot generalization. Second, our \textbf{Targeted Manifold Scattering (TMS)} objective uses preference-based mining to locally scatter the forget domain's stylistic geometry. Evaluated across PACS, OfficeHome, and DomainNet, our method vastly improves open-vocabulary generalization over existing baselines. Crucially, it delivers exceptional sample efficiency, outperforming peak 8-shot baseline results with only 4 shots.
\end{abstract}

\section{Introduction}

Pre-trained Vision-Language Models (VLMs) \cite{radford2021learning, faghri2025mobileclip2} exhibit remarkable zero-shot generalization, enabling them to recognize a vast array of open-vocabulary concepts without additional training. However, this strong generalization also means VLMs encode and retain information that may be unnecessary, computationally wasteful, or hazardous in specific downstream applications \cite{wang2021variational, shokri2017membership}. To address this, approximate machine unlearning has emerged to selectively remove specific knowledge while preserving overall model utility \cite{bourtoule2021machine, graves2021amnesiac}. Historically, approximate unlearning in VLMs has predominantly focused on \textit{class unlearning} wherein the task is to degrade recognition accuracy for specific object categories while maintaining accuracy for others \cite{golatkar2020eternal, fan2023salun, huang2024unified, kuwana2024black}. 

However, simply erasing an object class is often insufficient for real-world deployments when using pretrained VLMs. Consider an autonomous driving system: the model must reliably detect real physical cars to prevent collisions, but it must strictly avoid recognizing illustrated cars depicted on roadside advertisements as real vehicles \cite{kawamura2025approximate}. Standard class unlearning would blindly erase the concept of "car" entirely. This necessity has motivated the recent introduction of Approximate Domain Unlearning (ADU) \cite{kawamura2025approximate}, which aims to selectively reduce recognition accuracy for images from specified "forget" domains while preserving accuracy for "retain" domains. 

This requirement extends to high-stakes fields such as medical AI. Clinical VLMs pre-trained on internet-scale data often ingest "textbook diagrams" alongside real radiological scans \cite{tu2024towards}. Because diagrams contain artificial shortcuts such as idealized geometries, color-coding, and bright arrows, models can learn spurious correlations \cite{geirhos2020shortcut, degrave2021ai}, leading to hallucinated diagnoses when deployed on noisy, real-world patient X-rays. To ensure safety, the "diagram" domain must be entirely unlearned. 

Despite the clear practical motivation for domain unlearning, we identify a fundamental flaw in its current formulation and evaluation. Existing ADU methodologies typically evaluate unlearning performance on the exact same set of object classes that were presented during the unlearning fine-tuning phase \cite{kawamura2025approximate}. This closed-vocabulary protocol masks a critical failure mode: real-world environments are inherently open-ended. If an autonomous system is fine-tuned to unlearn illustrations using only common traffic classes, it will fail to generalize that domain erasure when encountering a billboard featuring a novel concept, such as a cartoon dinosaur. Similarly, in the medical domain, it is practically impossible to provide an unlearning algorithm with diagrammatic examples of every known rare disease during fine-tuning. If a domain erasure algorithm entangles stylistic features with the specific semantic classes seen during training, the VLM loses its foundational zero-shot ability to decouple pedagogical style from medical semantics on unseen, rare conditions.
\begin{figure*}[t]
    \centering
    \includegraphics[height=6cm, width=\linewidth, keepaspectratio]{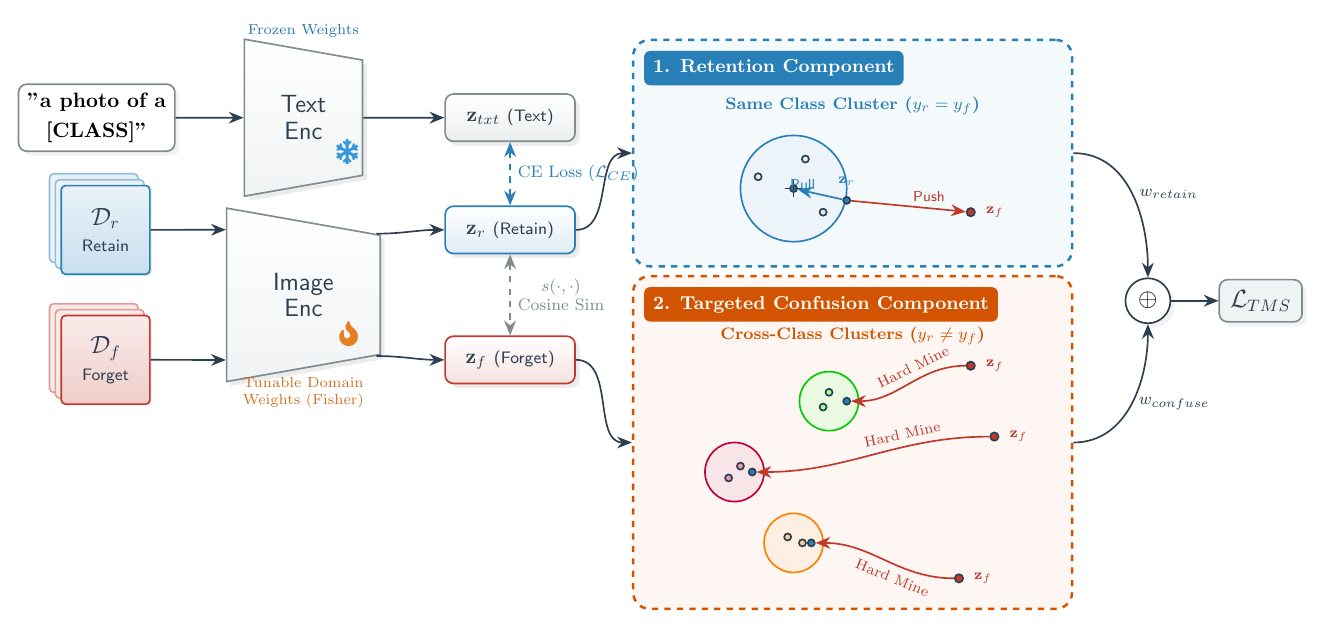}
    \caption{\textbf{Overview of the Targeted Manifold Scattering (TMS) framework.} Images from the retain ($\mathcal{D}_r$) and forget ($\mathcal{D}_f$) domains are processed through a Fisher-masked Image Encoder to isolate domain-specific updates. The TMS loss applies two decoupled geometric forces: (1) The \textbf{Retention Component} protects semantic integrity by contrasting same-class retain and forget embeddings ($y_r = y_f$). (2) The \textbf{Targeted Confusion Component} actively diffuses the forget representations by hard-mining cross-class retain clusters ($y_r \neq y_f$) and scattering the forget embeddings ($z_f$) into these unrelated regions, achieving efficient erasure without global semantic collapse.}
    \label{fig:TMS_architecture}
\end{figure*}
Furthermore, we demonstrate mathematically that current geometric regularizers exacerbate this catastrophic forgetting. Methods relying on maximizing the Maximum Mean Discrepancy (MMD) \cite{gretton2012kernel} to explicitly separate domain distributions in the latent space inadvertently compute a global displacement vector heavily biased by the semantic centroids of the seen unlearning classes. When applied via prompt tuning \cite{kawamura2025approximate, zhou2022learning}, MMD applies a global affine shift that misaligns and destroys the zero-shot generalization manifold for unseen classes. Similarly, rigid geometric regularizers like the hard-margin Triplet loss \cite{hoffer2015deep} apply step-function gradients that treat all violations equally once a threshold is crossed. This induces gradient oscillation and promotes \textit{manifold translation} wherein the forget domain is relocated into an isolated space rather than truly erasing its stylistic structure via scattering the representations locally.

To overcome these fundamental limitations, we formalize \textbf{Open-Vocabulary Domain Unlearning (OVDU)}. We argue that true domain unlearning must be class-agnostic, enabling a model to unlearn a domain's stylistic properties using only a small proxy subset of classes ($\mathcal{C}_{train}$) and successfully generalizing that erasure to entirely unseen concepts ($\mathcal{C}_{test}$). To achieve this, we build upon Fisher-guided unlearning principles \cite{golatkar2020eternal, liu2023unlearning}, we demonstrate that identifying and isolating parameters selectively sensitive to the forget domain projects gradient updates orthogonally to the semantic subspace. This provides a critical mathematical safeguard, insulating unseen classes from unlearning interference. To achieve true domain erasure, we introduce \textbf{Targeted Manifold Scattering (TMS)}. Unlike MMD or hard hinge based Triplet Loss, TMS utilizes a graduated log-sigmoid formulation coupled with hard mining to force forget embeddings to actively mimic unrelated cross-class retain embeddings. This induces bounded, localized \textit{manifold diffusion}, effectively shattering the stylistic geometry of the forget domain without disturbing global semantic alignment.

Our main contributions are summarized as follows:\\

\begin{itemize*}
    \item \textbf{A Novel Problem Setting and Protocol:} We identify the data leakage in current domain unlearning literature and introduce the Open-Vocabulary Domain Unlearning (OVDU) protocol, which rigorously evaluates zero-shot generalization by separating unlearning classes ($\mathcal{C}_{train}$) from testing classes ($\mathcal{C}_{test}$).\\
    \item \textbf{Targeted Manifold Scattering via Parameter Isolation:} As our primary contribution, we introduce a continuous geometric objective, Targeted Manifold Scattering, which decouples retention and targeted confusion forces to actively scatter the stylistic geometry of the forget domain. By restricting gradient flow exclusively to domain-sensitive weights using Fisher Information Mask, we mathematically prevent semantic collapse on unseen classes, enabling our framework to vastly outperform existing baselines.\\
    \item \textbf{Few-shot Sample Efficiency:} Extensive evaluations across the three different domain generalization benchmarks validates that our framework decisively outperforms existing baselines under the rigorous OVDU protocol. Beyond superior zero-shot domain erasure on held-out classes, our approach yields exceptional sample efficiency, nearly halving the data requirements of prior methods.
\end{itemize*}

\section{Related Work}

\begin{figure}[t] 
    \centering
    \resizebox{\columnwidth}{!}{\includegraphics[height=4.5cm, keepaspectratio]{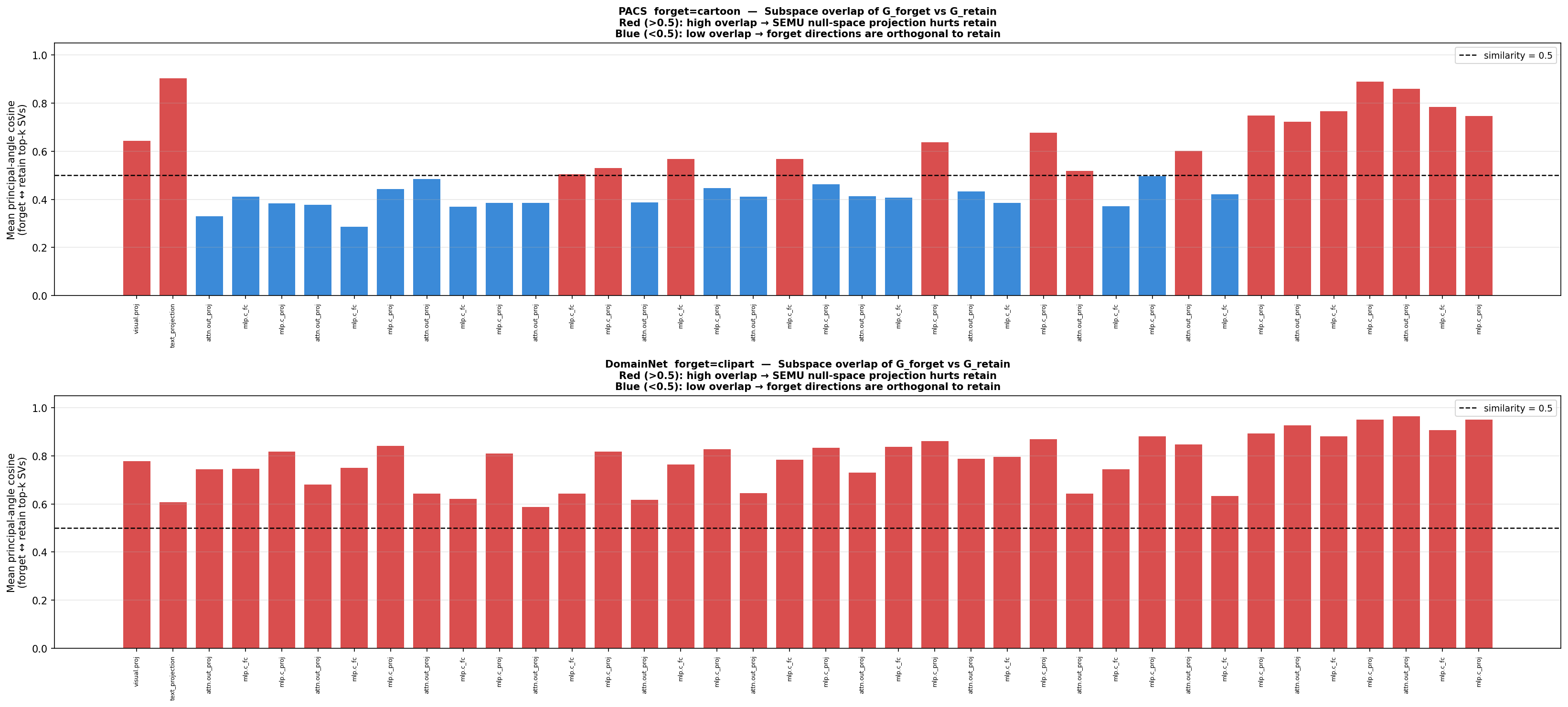}}
    
    \vspace{-12pt} 
    \caption{\small \textbf{Domain subspace entanglement across vision encoder layers.} This figure quantifies the structural overlap between domains in pre-trained VLMs. The x-axis represents the progressive layers of the vision encoder, while the y-axis measures the mean principal angle cosine, which specifically captures the overlap between the top $k$-singular vectors of the extracted domain subspaces (Forget and Retain). The prominent red bars denote exceptionally high structural similarity between the targeted forget domain and the retain domains across nearly all layers.}
    \label{fig:domain_subspaces}
    \vspace{-15pt} 
\end{figure}

\textbf{Machine Unlearning.} Traditional vision unlearning predominantly focuses on \textit{class erasure} via parameter isolation or weight editing (e.g., Task Arithmetic \cite{ilharco2022editing}, NegMerge \cite{kim2025negmerge}, or Fisher Information masks \cite{liu2023unlearning}). However, these methods struggle with \textit{domain unlearning}, where the objective shifts from deleting discrete concepts to decoupling pervasive, highly entangled stylistic renderings from the latent space (as illustrated in Figure~\ref{fig:domain_subspaces}).

\textbf{Approximate Domain Unlearning (ADU).} Recent ADU frameworks \cite{kawamura2025approximate} address domain erasure using prompt tuning and Maximum Mean Discrepancy (MMD) \cite{gretton2012kernel} for global feature separation. Yet, these methods rely on a closed-vocabulary assumption. In open-vocabulary scenarios, prompt-based MMD induces a severe global affine shift that degrades zero-shot generalization. Conversely, our approach abandons prompt tuning, leveraging Fisher-guided surgical parameter editing to isolate and degrade stylistic weights without disrupting foundational semantics.

\textbf{Geometric Regularization.} Rigid metric learning objectives (e.g., Triplet Loss \cite{hoffer2015deep}) typically cause simple \textit{manifold translation} rather than true feature disentanglement. Inspired by Direct Preference Optimization (DPO) \cite{rafailov2023direct} and Negative Preference Optimization (NPO) \cite{zhang2024negative}, we introduce Targeted Manifold Scattering (TMS). TMS transitions preference optimization to continuous geometric editing, using hard mining to locally shatter the forget domain's stylistic geometry while strictly preserving cross-class generalization. 

\textit{For a comprehensive discussion of related literature, extended theoretical comparisons, and unlearning mechanics, please refer to Appendix~\ref{app:extended_related_work}.}

\section{Methodology}

\subsection{Problem Formulation: Open-Vocabulary Domain Unlearning}

Let dataset $\mathcal{D} = \{(x_i, y_i, d_i)\}_{i=1}^N$ comprise inputs $x$, class labels $y \in \mathcal{C}$, and domain labels $d \in \{\mathcal{D}_f, \mathcal{D}_r\}$, where $\mathcal{D}_f$ and $\mathcal{D}_r$ denote the forget and retain domains, respectively. Given a pre-trained VLM parameterized by $\theta$, domain unlearning yields $\theta_{\mathrm{unlearned}}$ that severely degrades recognition on $\mathcal{D}_f$ while strictly preserving $\mathcal{D}_r$. Standard Approximate Domain Unlearning (ADU) \cite{kawamura2025approximate} assumes a closed-vocabulary setting where the class availability fraction exposed during fine-tuning is $\gamma = |\mathcal{C}_{\mathrm{train}}|/|\mathcal{C}| = 1.0$. To evaluate true stylistic decoupling, we introduce the \textbf{Open-Vocabulary Domain Unlearning (OVDU)} protocol, partitioning $\mathcal{C}$ into mutually exclusive seen classes ($\mathcal{C}_{\mathrm{seen}}$, used strictly for training) and held-out unseen classes ($\mathcal{C}_{\mathrm{unseen}}$). 

To systematically test open-vocabulary robustness, we vary $\gamma = |\mathcal{C}_{\mathrm{seen}}|/|\mathcal{C}| \in \{0.25, 0.50, 0.75, 1.0\}$. Crucially, OVDU evaluates on the global space $\mathcal{C}_{\mathrm{test}} = \mathcal{C}_{\mathrm{seen}} \cup \mathcal{C}_{\mathrm{unseen}} = \mathcal{C}$ to verify that domain erasure is class-agnostic. For any evaluation sample $(x_{\mathrm{test}}, y_{\mathrm{test}}) \sim \mathcal{C}_{\mathrm{test}}$, the model must simultaneously achieve \textbf{(1) Global Retention:} strictly preserving structural alignment and accuracy for $x_{\mathrm{test}} \in \mathcal{D}_r$ regardless of whether $y_{\mathrm{test}}$ is seen or unseen, and \textbf{(2) Global Erasure:} uniformly failing to recognize the stylistic rendering of $x_{\mathrm{test}} \in \mathcal{D}_f$ across the entire test space $\mathcal{C}_{\mathrm{test}}$. Achieving this decoupling requires preventing unlearning gradients from entangling with the semantic basis vectors of $\mathcal{C}_{\mathrm{seen}}$. To mathematically guarantee this global semantic preservation, we introduce a parameter isolation strategy baseline governed by empirical Fisher Information.

\subsection{Semantic Preservation via Fisher-Guided Parameter Isolation}

To rigorously evaluate the necessity of surgical weight editing over global prompt tuning, we first formalize a foundational parameter-isolation baseline: Gradient Ascent (NegGrad) \cite{graves2021amnesiac} governed by an empirical Fisher Information Mask \cite{liu2023unlearning}. An unconstrained NegGrad objective aims to induce forgetting by maximizing the cross-entropy loss on the forget domain $\mathcal{D}_f$ while minimizing it on the retain domain $\mathcal{D}_r$:
\begin{equation}
g_i^{\mathrm{NegGrad}} = \nabla_{\theta_i} \mathcal{L}_{\mathrm{CE}}(x_r, y_r; \theta) - \lambda \nabla_{\theta_i} \mathcal{L}_{\mathrm{CE}}(x_f, y_f; \theta)
\end{equation}
where $(x_r, y_r) \in \mathcal{D}_r$, $(x_f, y_f) \in \mathcal{D}_f$, and $\lambda$ is a scaling hyperparameter. While this unconstrained gradient difference effectively erases domain recognition, it indiscriminately penalizes shared semantic weights, completely destroying the pre-trained VLM's zero-shot generalization capabilities on unseen classes of the retain domains \cite{kumar2022fine}.

To prevent this semantic collapse, parameter isolation is strictly required to decouple the weights responsible for stylistic domain rendering from those encoding generalized semantics. Let $\ell_j(\theta)$ denote the per-sample loss for a given instance $(x_j, y_j) \in \mathcal{D}$. We compute the diagonal empirical Fisher Information Matrix by aggregating the squared gradients of this per-sample loss over the $N$ samples in the dataset:
\begin{equation}
F_{\mathcal{D}}(i) = \frac{1}{N} \sum_{j=1}^{N} \left( \nabla_{\theta_i} \ell_j(\theta) \right)^2
\end{equation}
Crucially, we utilize the average of the squared gradients across the batch samples rather than the maximum. This design choice ensures that the resulting sensitivity scores capture a stable, domain-wide parameter importance distribution without being disproportionately skewed by individual outlier samples or dominant object classes. We then quantify domain sensitivity for each parameter $\theta_i$ using the ratio:
\begin{equation}
\rho_i = \frac{F_{\mathcal{D}_f}(i) - F_{\mathcal{D}_r}(i)}{F_{\mathcal{D}_f}(i) + F_{\mathcal{D}_r}(i) + \epsilon}
\end{equation}
where $\epsilon$ is set to a small value to ensure numerical stability.

This normalization elegantly handles domain unlearning because the forget dataset $\mathcal{D}_f$ and retain dataset $\mathcal{D}_r$ share the underlying semantic classes ($\mathcal{C}_{\mathrm{train}}$). Parameters encoding fundamental semantics remain active in both domains, yielding $\rho_i \le 0$, while parameters strictly dictating stylistic artifacts dominate in the forget domain, driving $\rho_i \to 1$. We isolate these stylistic weights via a binary mask with a predefined sparsity threshold $\tau$:
\begin{equation}
m_i = \mathbbm{1}[\rho_i > \tau] \in \{0, 1\}
\end{equation}
Routing NegGrad updates strictly through this mask ($\theta_i \leftarrow \theta_i - \eta (g_i^{\mathrm{NegGrad}} \cdot m_i)$) forces the unlearning gradients into the null space of the semantic feature extractor. This mathematically guarantees the preservation of zero-shot knowledge for unseen classes, establishing a rigorous baseline against which we compare our primary geometric objective.

\subsection{Targeted Manifold Scattering: Scattering the Domain Geometry}

Although the Fisher-guided NegGrad baseline provides a structural constraint to help preserve foundational knowledge, our empirical evaluations reveal a critical limitation in highly constrained, open-vocabulary regimes. Specifically, at low class fractions ($\gamma \le 0.50$), standard Gradient Ascent lacks a bounded, structural destination in the latent space. Because it simply pushes forget embeddings away by flattening the local loss landscape, the optimizer finds a "lazy" local minimum. Confronted with the strict bottleneck of the Fisher mask and a sparse gradient signal from the limited seen classes ($\mathcal{C}_{\mathrm{seen}}$), the model makes minimal perturbations to the domain weights, applying just enough adjustment to push the seen classes over the decision boundary. Consequently, the underlying structural geometry of the domain remains largely intact, resulting in weak unlearning on the seen classes and a complete failure to generalize this erasure to unseen classes ($\mathcal{C}_{\mathrm{unseen}}$).

Conversely, baseline methods like ADU rely on Maximum Mean Discrepancy (MMD) to explicitly separate domain distributions via prompt tuning. However, at low $\gamma$, MMD computes a separation shift based entirely on the empirical means of the small subset of seen classes. This calculated global shift, $\Delta \propto \mu_{\mathrm{f, seen}} - \mu_{\mathrm{r, seen}}$, becomes intrinsically entangled with the specific semantic geometry of $\mathcal{C}_{seen}$ rather than representing a pure domain translation. When this biased affine shift is applied globally across the latent space, an unseen retain embedding $z_r \in \mathcal{C}_{\mathrm{unseen}}$ is forcefully displaced ($z_r \to z_r + \Delta$). This implicitly drags the embedding out of its ground-truth semantic cluster, mathematically ensuring the catastrophic forgetting that severely degrades the model's performance even on the retain domain $\mathcal{D}_r$.

These contrasting failure modes illustrate that domain unlearning requires \textit{localized manifold diffusion}. Instead of indiscriminately pushing forget embeddings to infinity (NegGrad) or shifting the entire space globally (ADU), we must actively scatter the stylistic representations of the forget domain into dense, unrelated regions of the retain space. To achieve this, we introduce Targeted Manifold Scattering (TMS), a continuous geometric objective taking inspiration from Direct Preference Optimization (DPO) \cite{rafailov2023direct}. While standard representation scattering often relies on contrastive loss functions with rigid hinge margins, which impose discontinuous and uniform penalties that can easily destabilize delicate semantic boundaries, our objective leverages DPO's continuous log-sigmoid formulation. This ensures that the loss landscape remains smooth, yielding graduated gradients strictly proportional to the magnitude of the geometric violation and providing highly stable structural updates.

Let $z_r$ and $z_f$ denote the L2-normalized visual embeddings for the retain and forget domains, respectively, with corresponding labels $y_r$ and $y_f$. As overviewed in Figure \ref{fig:TMS_architecture}, rather than treating all forget embeddings as uniform negatives, our TMS objective induces structural tension using two decoupled rules:

\textbf{1. The Retention Component ($\mathcal{L}_{\mathrm{retain}}$):} This term protects the retain clusters from being encroached upon by same-class forget data. An anchor retain embedding $z_r$ is forced to explicitly prefer its own same-class retain positive ($z_{r\_pos}$) over a forget embedding $z_f$ of the same class. 
\begin{equation}
\mathcal{L}_{\mathrm{retain}} = \mathbb{E}_{(z_r, z_f) \mid y_r = y_f} \left[ -\log \sigma \left( \beta \left( s(z_r, z_{\mathrm{r\_pos}}) - s(z_r, z_f) - m_{\mathrm{retain}} \right) \right) \right]
\end{equation}
where $s(\cdot, \cdot)$ denotes cosine similarity, $\beta$ is the temperature scaling, and $m_{\mathrm{retain}}$ is the retention margin gap.

\textbf{2. The Targeted Confusion Component ($\mathcal{L}_{\mathrm{confuse}}$):} To ensure the model structurally dismantles the domain rather than finding a lazy local minimum, we force forget embeddings to actively hide within unrelated classes. Utilizing hard mining, each forget embedding $z_f$ dynamically seeks out the single cross-class retain embedding it is currently most similar to (i.e., the easiest to mistake). The model is explicitly penalized unless it pushes this cross-class similarity above a high confusion margin:
\begin{equation}
\mathcal{L}_{\mathrm{confuse}} = \mathbb{E}_{z_f} \left[ -\log \sigma \left( \beta \left( \max_{z_r \mid y_r \neq y_f} s(z_f, z_r) - m_{\mathrm{confuse}} \right) \right) \right]
\end{equation}

The final continuous geometric preference loss is a weighted sum of these decoupled forces:
\begin{equation}
\mathcal{L}_{\mathrm{TMS}} = w_{\mathrm{retain}} \mathcal{L}_{\mathrm{retain}} + w_{\mathrm{confuse}} \mathcal{L}_{\mathrm{confuse}}
\end{equation}

This preference objective is integrated directly with the base NegGrad cross-entropy (CE) terms using a scaling weight $\lambda_{pref}$:
\begin{equation}
\mathcal{L}_{\mathrm{total}} = (1 - \lambda_{\mathrm{ce}}) \mathcal{L}_{\mathrm{retain\_CE}} - \lambda_{\mathrm{ce}} \mathcal{L}_{\mathrm{forget\_CE}} + \lambda_{\mathrm{pref}} \mathcal{L}_{\mathrm{TMS}}
\end{equation}

The synergy between $\mathcal{L}_{TMS}$ and the Fisher mask is the core driver of our open-vocabulary domain unlearning improvements. Because the hard-mining confusion term demands deep structural reconfiguration to make a forget embedding mimic an entirely different semantic class, the optimizer cannot bypass the loss with superficial tweaks. Forced to rely strictly on the domain-sensitive weights isolated by the Fisher mask, the model systematically scatters the foundational representations rendering the forget domain. This bounded, localized diffusion comprehensively scatters the domain's geometry across $\mathcal{C}_{\mathrm{seen}}$, a structural disruption that cascades effortlessly to achieve robust, global erasure across the entire evaluation space ($\mathcal{C}_{\mathrm{seen}} \cup \mathcal{C}_{\mathrm{unseen}}$), while mathematically preventing the catastrophic semantic shift observed in global MMD methods. 

Furthermore, this extreme geometric demand renders our framework exceptionally sample-efficient. Because the hard-mining objective extracts maximal structural unlearning signal from every individual image, TMS effectively eliminates the reliance on massive, statistically stable datasets typically required to compute distribution shifts. Consequently, this targeted approach explicitly unlocks rapid, high-performance few-shot domain unlearning, achieving profound erasure with only a fraction of the data exposure required by prior works.

\section{Experiments}

\subsection{Experimental Setup}

\textbf{Datasets.} We extensively evaluate our method across three standard benchmarks, each featuring four visually distinct domains: \textbf{Office-Home}~\cite{venkateswara2017deep} (65 classes; Art, Clipart, Product, Real-World), \textbf{Mini-DomainNet}~\cite{peng2019moment} (126 classes; Clipart, Painting, Real, Sketch), and \textbf{PACS}~\cite{li2017deeper} (7 classes with severe stylistic shifts; Photo, Art Painting, Cartoon, Sketch).

\textbf{Baselines.} We benchmark our proposed Targeted Manifold Scattering (TMS) against two primary methods: 
1) \textbf{ADU} \cite{kawamura2025approximate}, a recent state-of-the-art global prompt-tuning approach for domain unlearning, and 
2) \textbf{NegGrad+Fisher}, the rigorous parameter-isolation baseline formalized in Section 3.1, which applies empirical Fisher-guided gradient ascent directly to the model weights. Additional baselines such as  SEMU \cite{sendera2025semu}, and SAM based unlearning \cite{tang2025sharpness} are considered in the Appendix. 

\textbf{Evaluation Protocol.} Following the setup defined in Section~3.1, we adopt a strict few-shot ($K=8$) unlearning protocol using a class fraction $\gamma \in \{0.25, 0.50, 0.75, 1.0\}$. We evaluate on all classes ($\mathcal{C}_{seen} \cup \mathcal{C}_{unseen}$) to assess both domain erasure and semantic retention, and further benchmark zero-shot generalization robustness on ImageNet-1K~\cite{deng2009imagenet}.

\textbf{Implementation Details.} We employ OpenCLIP (ViT-B/16~\cite{dosovitskiy2020image}) with a frozen text encoder, updating only the visual weights constrained by a Fisher sensitivity mask ($\tau=0.3$). Our TMS loss is applied to the pre-projection CLS-token space to ensure foundational structural updates. To prevent gradient imbalance during few-shot sampling, the retain and forget cross-entropy terms are dynamically scaled by their sample proportions ($\alpha_r = \frac{n_r}{n_r+n_f}$ and $\alpha_f = \frac{n_f}{n_r+n_f}$). Models are optimized using AdamW~\cite{loshchilov2017decoupled} with a learning rate of 1e-5 for 10 epochs. The network is optimized using AdamW \cite{loshchilov2017decoupled} with a learning rate of 1e-5 for 10 epochs. Additional details are provided in the Appendix.

\textbf{Evaluation Metrics.} We evaluate unlearning performance using \textbf{Retain Accuracy ($\mathcal{A}_r$)} (higher is better) and \textbf{Forget Accuracy ($\mathcal{A}_f$)} (lower is better). To quantify the overall trade-off between effective domain erasure and zero-shot semantic preservation, we report their \textbf{Harmonic Mean ($\mathcal{H}$)}. Specifically, $\mathcal{H}$ is computed between $\mathcal{A}_r$ and the erasure rate ($\mathcal{E} = 100 - \mathcal{A}_f$) as $\mathcal{H} = \frac{2 \cdot \mathcal{A}_r \cdot \mathcal{E}}{\mathcal{A}_r + \mathcal{E}}$. All reported results are averaged across multiple independent runs to ensure statistical robustness.

\begin{table*}[htbp]
    \centering
    \caption{Unlearning performance across different class fractions ($\gamma$) with $|D_f| = 1$. \textbf{F}: Forget domain accuracy (\%), \textbf{R}: Retain domain accuracy (\%), \textbf{HM}: Harmonic Mean (\%). Best values for each fraction and dataset are in \textbf{bold}.}
    \label{tab:class_fractions}
    \resizebox{\textwidth}{!}{
    \begin{tabular}{ll ccc ccc ccc ccc}
        \toprule
        \multirow{2}{*}{\textbf{Dataset}} & \multirow{2}{*}{\textbf{Method}} & \multicolumn{3}{c}{$\mathbf{\gamma = 0.25}$} & \multicolumn{3}{c}{$\mathbf{\gamma = 0.50}$} & \multicolumn{3}{c}{$\mathbf{\gamma = 0.75}$} & \multicolumn{3}{c}{$\mathbf{\gamma = 1.0}$} \\
        \cmidrule(lr){3-5} \cmidrule(lr){6-8} \cmidrule(lr){9-11} \cmidrule(lr){12-14}
        & & F $\downarrow$ & R $\uparrow$ & HM $\uparrow$ & F $\downarrow$ & R $\uparrow$ & HM $\uparrow$ & F $\downarrow$ & R $\uparrow$ & HM $\uparrow$ & F $\downarrow$ & R $\uparrow$ & HM $\uparrow$ \\
        \midrule
        \multirow{3}{*}{\textbf{OfficeHome}}
        & Baseline   & 59.54 & \textbf{65.65} & 50.07 & 51.62 & 67.92 & 56.51 & 36.97 & \textbf{71.90} & 67.17 & 31.31 & 77.90 & 73.01 \\
        & ADU        & \textbf{21.97} & 27.61 & 40.79 & \textbf{28.26} & 49.22 & 58.38 & 33.63 & 67.96 & 67.16 & 36.37 & \textbf{80.36} & 71.02 \\
        & Ours       & 53.33 & 63.24 & \textbf{53.71} & 39.34 & \textbf{68.33} & \textbf{64.27} & \textbf{24.90} & 68.68 & \textbf{71.75} & \textbf{24.61} & 75.45 & \textbf{75.42} \\
        \midrule
        \multirow{3}{*}{\textbf{DomainNet}}
        & Baseline   & 55.58 & \textbf{77.72} & 56.53 & 27.98 & \textbf{77.08} & 74.46 & 20.04 & \textbf{77.24} & 78.58 & 18.02 & 77.77 & 79.82 \\
        & ADU        & \textbf{14.64} & 24.45 & 38.01 & \textbf{19.68} & 49.25 & 61.06 & 23.41 & 67.27 & 71.63 & 27.86 & \textbf{81.48} & 76.53 \\
        & Ours       & 34.04 & 74.97 & \textbf{70.18} & 20.06 & 75.39 & \textbf{77.60} & \textbf{14.60} & 75.97 & \textbf{80.41} & \textbf{14.44} & 76.84 & \textbf{80.97} \\
        \midrule
        \multirow{3}{*}{\textbf{PACS}}
        & Baseline   & 78.64 & 90.02 & 34.52 & 57.75 & \textbf{93.76} & 58.25 & 59.40 & \textbf{93.42} & 56.60 & 41.01 & 95.60 & 72.96 \\
        & ADU        & \textbf{64.97} & 28.65 & 31.52 & 64.07 & 59.53 & 44.81 & 68.36 & 74.40 & 44.40 & 72.48 & 93.16 & 42.49 \\
        & Ours       & 76.085 & \textbf{90.39} & \textbf{37.82} & \textbf{47.06} & 93.49 & \textbf{67.60} & \textbf{47.07} & 93.24 & \textbf{67.53} & \textbf{38.35} & \textbf{95.64} & \textbf{74.97} \\
        \bottomrule
    \end{tabular}
    }
\end{table*}

\begin{table*}[htbp]
    \centering
    \caption{Ablation study on the DomainNet dataset comparing the individual components of our TMS loss formulation (Confuse-only, Retain-only, and Both). \textbf{F}: Forget domain accuracy (\%), \textbf{R}: Retain domain accuracy (\%), \textbf{HM}: Harmonic Mean (\%). Best aggregate values are in \textbf{bold}.}
    \label{tab:ablation_loss}
    \resizebox{\textwidth}{!}{
    \begin{tabular}{lc *{9}{>{\centering\arraybackslash}p{1.2cm}}}
        \toprule
        \multirow{2}{*}{\textbf{Domain}} & \multirow{2}{*}{\textbf{$\gamma$}} & \multicolumn{3}{c}{\textbf{Confuse Only}} & \multicolumn{3}{c}{\textbf{Retain Only}} & \multicolumn{3}{c}{\textbf{Both Losses (Ours)}} \\
        \cmidrule(lr){3-5} \cmidrule(lr){6-8} \cmidrule(lr){9-11}
        & & F $\downarrow$ & R $\uparrow$ & HM $\uparrow$ & F $\downarrow$ & R $\uparrow$ & HM $\uparrow$ & F $\downarrow$ & R $\uparrow$ & HM $\uparrow$ \\
        \midrule
        \multirow{2}{*}{\textbf{Clipart}} 
        & 0.50 & 19.68 & 77.25 & 78.75 & 20.63 & \textbf{78.04} & 78.70 & \textbf{17.93} & 76.82 & \textbf{79.35} \\
        & 1.00 & \textbf{16.03} & \textbf{78.10} & \textbf{80.93} & 18.73 & 77.78 & 79.49 & 17.30 & 77.08 & 79.79 \\
        \midrule
        \multirow{2}{*}{\textbf{Painting}} 
        & 0.50 & 21.27 & 78.15 & 78.44 & 26.03 & \textbf{80.37} & 77.04 & \textbf{17.46} & 76.82 & \textbf{79.57} \\
        & 1.00 & 17.78 & 78.94 & 80.55 & 23.33 & \textbf{80.37} & 78.47 & \textbf{17.30} & 78.88 & \textbf{80.74} \\
        \midrule
        \multirow{2}{*}{\textbf{Real}} 
        & 0.50 & \textbf{31.27} & 71.48 & \textbf{70.08} & 38.57 & \textbf{72.86} & 66.66 & 32.38 & 72.70 & 70.06 \\
        & 1.00 & 23.81 & \textbf{76.61} & 76.40 & 18.73 & 74.23 & 77.59 & \textbf{13.65} & 73.75 & \textbf{79.55} \\
        \midrule
        \multirow{2}{*}{\textbf{Sketch}} 
        & 0.50 & 14.44 & 75.08 & 79.98 & 13.49 & \textbf{76.98} & \textbf{81.47} & \textbf{12.53} & 75.23 & 80.88 \\
        & 1.00 & 11.59 & 76.46 & 82.00 & 10.79 & \textbf{77.99} & 83.22 & \textbf{9.52} & 77.77 & \textbf{83.64} \\
        \midrule
        \midrule
        \multirow{2}{*}{\textbf{DomainNet (Avg)}} 
        & 0.50 & 21.67 & 75.49 & 76.81 & 24.68 & \textbf{77.06} & 75.97 & \textbf{20.06} & 75.39 & \textbf{77.60} \\
        & 1.00 & 17.30 & 77.53 & 79.97 & 17.90 & \textbf{77.59} & 79.69 & \textbf{14.44} & 76.84 & \textbf{80.97} \\
        \bottomrule
    \end{tabular}
    }
\end{table*}

\begin{table*}[htbp]
    \centering
    \caption{Unlearning performance on DomainNet for multiple domain forgetting ($|D_f| = 3$) across different class fractions ($\gamma$). \textbf{F}: Forget domain accuracy (\%), \textbf{R}: Retain domain accuracy (\%), \textbf{HM}: Harmonic Mean (\%). Best values for each fraction are in \textbf{bold}.}
    \label{tab:multiple_forgetting}
    \resizebox{\textwidth}{!}{
    \begin{tabular}{ll ccc ccc ccc ccc}
        \toprule
        \multirow{2}{*}{\textbf{Dataset}} & \multirow{2}{*}{\textbf{Method}} & \multicolumn{3}{c}{$\mathbf{\gamma = 0.25}$} & \multicolumn{3}{c}{$\mathbf{\gamma = 0.50}$} & \multicolumn{3}{c}{$\mathbf{\gamma = 0.75}$} & \multicolumn{3}{c}{$\mathbf{\gamma = 1.0}$} \\
        \cmidrule(lr){3-5} \cmidrule(lr){6-8} \cmidrule(lr){9-11} \cmidrule(lr){12-14}
        & & F $\downarrow$ & R $\uparrow$ & HM $\uparrow$ & F $\downarrow$ & R $\uparrow$ & HM $\uparrow$ & F $\downarrow$ & R $\uparrow$ & HM $\uparrow$ & F $\downarrow$ & R $\uparrow$ & HM $\uparrow$ \\
        \midrule
        \multirow{3}{*}{\textbf{DomainNet}}
        & Baseline & 8.92 & 59.87 & 72.25 & \textbf{4.54} & 61.06 & 74.48 & \textbf{3.26} & 62.13 & 75.67 & \textbf{3.33} & 63.45 & 76.61 \\
        & ADU      & 9.97 & 24.88 & 38.99 & 13.70 & 47.61 & 61.37 & 14.85 & 64.92 & 73.67 & 16.79 & \textbf{77.06} & \textbf{80.02} \\
        & Ours     & \textbf{7.11} & \textbf{60.19} & \textbf{73.05} & 5.07 & \textbf{62.45} & \textbf{75.34} & 4.81 & \textbf{66.10} & \textbf{78.02} & 4.59 & 67.21 & 78.86 \\
        \bottomrule
    \end{tabular}
    }
\end{table*}

\subsection{Experimental Results}

\textbf{Overall Efficacy and the TMS Formulation:} Table~\ref{tab:class_fractions} demonstrates that our TMS formulation strictly achieves the highest Harmonic Mean (HM) across all datasets (OfficeHome, DomainNet, and PACS) and class fractions ($\gamma$). This superior performance stems directly from our dual-objective design: a \textbf{confusion component} that aggressively unlearns target domain features, and a \textbf{retain component} that explicitly anchors preserved knowledge.

\textbf{Overcoming the Stability-Plasticity Dilemma:} The results reveal a severe trade-off in existing methods. While ADU achieves strong forgetting at low fractions ($\gamma \le 0.50$), it suffers catastrophic degradation on the retain set (e.g., DomainNet retain accuracy collapses to 24.45\% at $\gamma=0.25$, whereas ours maintains 74.97\%). Conversely, the Baseline preserves retain knowledge but fails to sufficiently unlearn the target concepts. By leveraging our independent retain and confusion components, our method successfully decouples these objectives, resolving the dilemma to deliver optimal HM.

\textbf{Robustness to Unlearning Scope ($\gamma$):} As the unlearning scope expands to the full domain ($\gamma=1.0$), our confusion component efficiently drives forget accuracies to the lowest (best) levels across all benchmarks (e.g., 14.44\% on DomainNet, 24.61\% on OfficeHome). Unlike ADU, which struggles with mode collapse at varying fractions, our formulation scales seamlessly, ensuring that the retain component stabilizes the network while the confusion component maximizes unlearning capacity.

\textbf{Multi-Domain Unlearning:} Table \ref{tab:multiple_forgetting} shows  that even in the multi-domain forgetting scenario, our TMS formulation demonstrates superior stability and balance compared to both the Baseline and ADU. While ADU suffers from a catastrophic collapse in retention at lower class fractions (e.g., 24.88\% at $\gamma=0.25$), our method maintains robust performance and achieves the highest Harmonic Mean for the majority of settings. These results confirm that the synergy between our retain and confusion components scales effectively to complex unlearning tasks, providing precise domain erasure while safeguarding foundational knowledge significantly better than existing approaches.

\begin{table}[htbp]
    \centering
    \caption{Zero-shot generalization robustness on ImageNet-val and CIFAR-100 after unlearning specific DomainNet domains. Higher values indicate better preservation of foundational, open-vocabulary knowledge without catastrophic semantic forgetting. The original base model serves as the upper bound reference.}
    \label{tab:imagenet_robustness}
    \begin{tabular}{ll cc}
        \toprule
        \textbf{Forget Domain} & \textbf{Method} & \textbf{ImageNet-val (\%)} $\uparrow$ & \textbf{CIFAR-100 (\%)} $\uparrow$ \\
        \midrule
        \multicolumn{2}{l}{\textit{Base Model (No Unlearning)}} & \textit{63.38} & \textit{61.99} \\
        \midrule
        \multirow{3}{*}{\textbf{Clipart}}
        & Baseline & 54.11 & 57.40 \\
        & ADU      & 36.73 & 49.37 \\
        & Ours     & \textbf{57.33} & \textbf{57.50} \\
        \midrule
        \multirow{3}{*}{\textbf{Painting}}
        & Baseline & \textbf{51.76} & 50.45 \\
        & ADU      & 28.45 & 32.54 \\
        & Ours     & 49.30 & \textbf{52.26} \\
        \midrule
        \multirow{3}{*}{\textbf{Sketch}}
        & Baseline & 46.87 & 52.49 \\
        & ADU      & 39.42 & \textbf{57.01} \\
        & Ours     & \textbf{48.82} & 56.21 \\
        \bottomrule
    \end{tabular}
\end{table}

\begin{figure}[t] 
    \centering
    \resizebox{\columnwidth}{!}{\includegraphics[height=3.5cm, keepaspectratio]{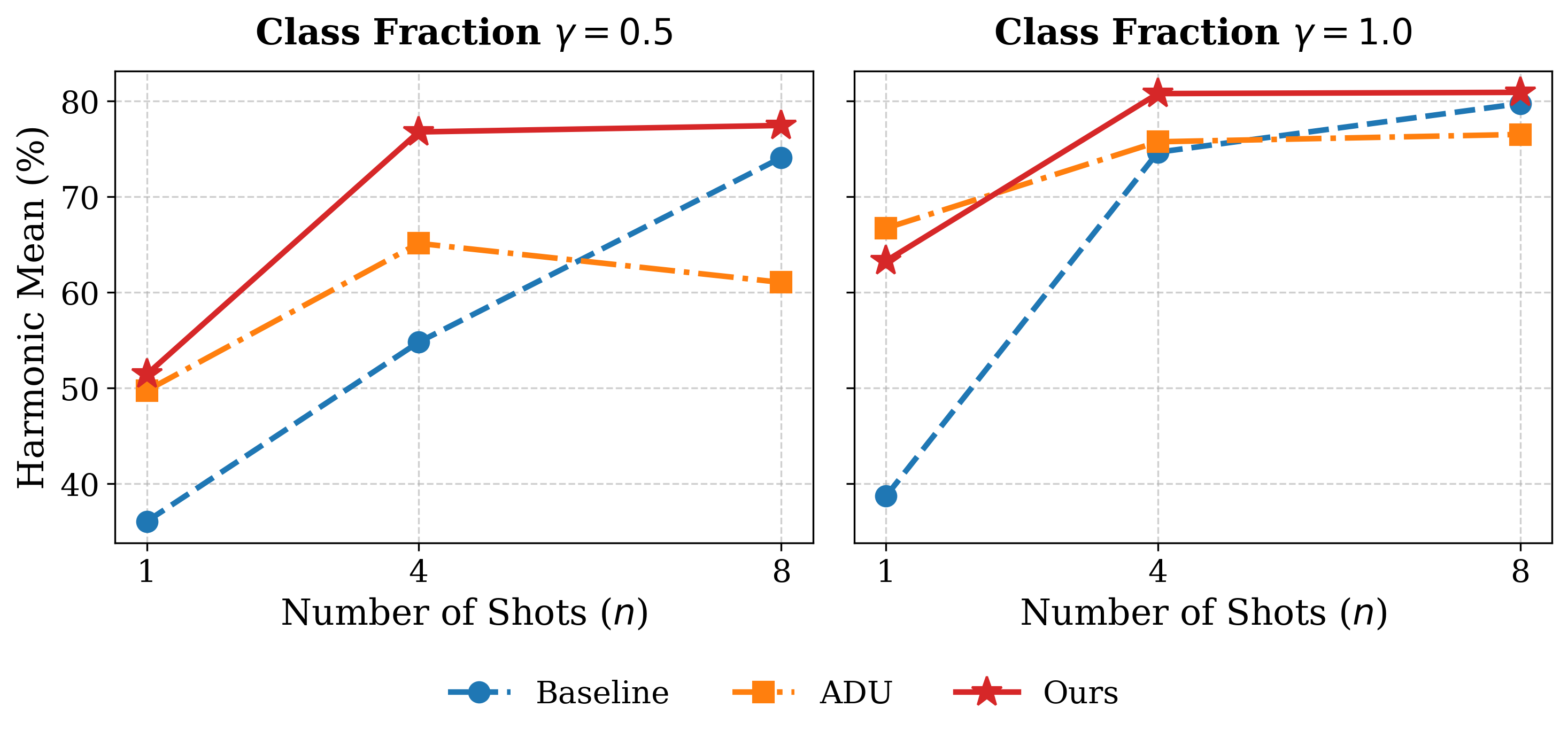}}
    
    \vspace{-12pt} 
    \caption{\small \textbf{Sensitivity to the Number of Training Samples.} Unlearning capacity (HM) compared across varying shots (n) per domain. While the Baseline and ADU show limited or unstable improvement in low-data regimes, our method demonstrates significantly faster convergence, achieving superior unlearning performance with just 4 shots.}
    \label{fig:sample_efficiency}
    \vspace{-15pt} 
\end{figure}

\textbf{Ablation Study: Loss Components.} Table~\ref{tab:ablation_loss} isolates the effects of the \textit{confuse} and \textit{retain} components within our TMS loss on the DomainNet benchmark. When applied individually, a clear stability-plasticity trade-off emerges. The \textit{confuse-only} loss generally promotes target forgetting but can destabilize retain accuracy, whereas the \textit{retain-only} loss maximizes knowledge preservation (achieving the highest average Retain accuracy of 77.59\% at $\gamma=1.0$) but struggles to induce sufficient forgetting. 

By unifying both objectives, our complete formulation explicitly resolves this trade-off. The synergistic combination of targeted domain confusion with explicit structural retention yields the lowest overall forget accuracies ($F=20.06\%$ and $14.44\%$) while maintaining highly competitive retention. Consequently, our joint formulation achieves the highest average Harmonic Mean ($\text{HM}=77.60\%$ and $80.97\%$) across both class fractions, confirming that both components are strictly necessary for optimal domain unlearning.

\textbf{Few-Shot Sample Efficiency:} Figure~\ref{fig:sample_efficiency} evaluates performance under extreme few-shot constraints ($n \in \{1, 4, 8\}$ shots per class). Our method demonstrates significantly faster convergence, achieving highly effective unlearning with just 4 shots and strictly outperforming the peak 8-shot performance of both the baseline and ADU. For closed-domain unlearning ($\gamma=1.0$), our 4-shot Harmonic Mean (80.80\%) surpasses the 8-shot baseline (79.76\%) and ADU (76.53\%). This trend holds for open-vocabulary domain unlearning ($\gamma=0.5$) as well, where our 4-shot accuracy (76.79\%) decisively beats the 8-shot baseline (74.09\%) and ADU (61.06\%). Unlike competing methods that degrade sharply in ultra-low data regimes ($n \le 4$), our approach maintains robust feature decoupling, drastically reducing the data curation bottleneck required for practical unlearning.

\textbf{Zero-Shot Generalization Robustness.} 
A critical risk when unlearning specific visual domains (e.g., Painting or Sketch) is \textit{catastrophic semantic forgetting}. To verify that our unlearning process is strictly localized and not damagingly broad, we benchmark zero-shot robustness on expansive, out-of-distribution datasets (ImageNet-val and CIFAR-100) following the domain erasure. As shown in Table~\ref{tab:imagenet_robustness}, our TMS formulation excels at preserving global semantic knowledge. While prior methods like ADU suffer severe foundational degradation (e.g., dropping ImageNet accuracy by roughly 35\% when unlearning the Painting domain), our method consistently anchors generalization performance near the original Base Model upper bound. Across all tested domains, our approach secures the highest overall retention on ImageNet and CIFAR-100 compared to the baselines. Crucially, this proves that direct parameter-tuning for unlearning—when formulated correctly—does not inherently destroy foundational generalization. Consequently, it demonstrates that relying on prompt-tuning-based unlearning is not strictly mandatory to safely avoid catastrophic forgetting in large vision-language models. Additional results can be found in the Appendix.

\section{Conclusion and Limitations}
We formalized the \textbf{Open-Vocabulary Domain Unlearning (OVDU)} protocol to address the hidden risk of catastrophic semantic forgetting in VLMs. By employing Fisher-guided parameter editing with our \textbf{Targeted Manifold Scattering (TMS)} objective, we demonstrate that domain stylistic geometry can be scattered without compromising foundational zero-shot intelligence. Our framework achieves superior stability-plasticity balance across 3 datasets, outperforming 8-shot baselines with only 4 samples. This confirms that constrained parameter-tuning is a robust and sample-efficient path for precise domain erasure. Despite these gains, several limitations remain. Our reliance on a diagonal Fisher Information Matrix approximation could be further refined by exploring off-diagonal parameter dependencies. Additionally, while we prove scalability to $|D_f|=3$, future work should investigate the upper bounds of multi-domain erasure and extend the TMS objective to the text encoder to prevent cross-modal information leakage.

\bibliographystyle{plainnat}
\bibliography{references}


\appendix
\section {Appendix}
\subsection{Extended Related Work}
\label{app:extended_related_work}

\subsubsection{Machine Unlearning: From Classes to Domains}
Machine unlearning aims to remove the influence of specific training data from a model without the prohibitive computational cost of retraining from scratch \cite{bourtoule2021machine, graves2021amnesiac}. In the context of vision models, the literature has predominantly focused on \textit{class unlearning} techniques designed to make the model forget specific object categories while preserving accuracy on retained classes \cite{fan2023salun, golatkar2020eternal, ilharco2022editing}. Within this realm of class unlearning, parameter isolation and direct weight editing have emerged as crucial strategies to prevent the catastrophic forgetting of retained knowledge. Recent advancements manipulate the parameter space directly to induce forgetting, employing techniques such as gradient weight negation \cite{graves2021amnesiac}, subtracting task-specific vectors via Task Arithmetic \cite{ilharco2022editing}, or applying consensual weight merging as seen in NegMerge \cite{kim2025negmerge}. Other approaches target specific representational subspaces, leveraging Singular Value Decomposition (SEMU) \cite{sendera2025semu} or gradient-based weight saliency \cite{fan2023salun} to mask and alter only a fraction of critical parameters. Similarly, Fisher-based approaches utilize empirical Fisher Information to identify and isolate the specific subset of weights that are disproportionately sensitive to the forget data \cite{liu2023unlearning}. 

While these advanced weight-editing and isolation techniques have proven highly effective at preserving semantic boundaries during class erasure, their potential has remained entirely unexplored in the context of domain unlearning. Here, the objective shifts fundamentally: rather than deleting a discrete concept, the model must decouple a pervasive stylistic rendering from an open-vocabulary semantic space. Furthermore, as illustrated in Figure~\ref{fig:domain_subspaces}, the latent feature space of pre-trained VLMs exhibits significant entanglement among domain distributions, in stark contrast to typically well-separated class distributions. This severe overlap makes the successful application of conventional class unlearning techniques \cite{kodge2024deep, sendera2025semu, feng2025fg} inherently challenging for domain unlearning.

Class-level erasure is often too blunt for practical applications, prompting the recent introduction of Approximate Domain Unlearning (ADU) \cite{kawamura2025approximate}. ADU aims to suppress a model's recognition capability for specific domains (e.g., illustrations) while retaining it for others (e.g., real photos). To address this, current ADU methods rely on prompt tuning combined with Maximum Mean Discrepancy (MMD) to explicitly disentangle domain distributions. However, as we demonstrate, these methods are evaluated in a closed-vocabulary setting where the unlearning and testing classes overlap perfectly. When applied to an open-vocabulary setting, prompt-based MMD induces a $\mathcal{C}_{train}$-biased global affine shift that severely degrades the zero-shot generalization manifold for unseen classes.

\subsubsection{Domain Generalization and Unlearning in VLMs}
While foundation models \cite{faghri2025mobileclip2, radford2021learning} exhibit profound domain generalization, their highly entangled latent spaces complicate unlearning. This often leads prior ADU works to rely on global separation techniques like MMD \cite{gretton2012kernel}. However, we observe that the \textit{internal parameter space} of the visual encoder remains highly structured. Instead of forcing global feature disentanglement, our work leverages a Fisher Information Mask \cite{liu2023unlearning} to isolate and degrade specific stylistic weights. This surgical parameter updating prevents the catastrophic semantic interference seen in global tuning approaches.

\subsubsection{Geometric Regularization for Unlearning}
Rigid metric learning objectives (e.g., Triplet Loss \cite{hoffer2015deep}) are ill-suited for unlearning, as their hard margins typically cause simple \textit{manifold translation} (relocating clusters) rather than actual feature disentanglement. To overcome this, we draw inspiration from Direct Preference Optimization (DPO) \cite{rafailov2023direct} and its recent unlearning extensions like NPO \cite{zhang2024negative}. We transition preference-based unlearning from token-level probabilities to continuous geometric representation editing via our \textit{Targeted Manifold Scattering} (TMS) objective. By using hard mining to force forget embeddings to actively mimic unrelated cross-class retain embeddings, TMS induces a localized \textit{manifold diffusion} that scatters the stylistic geometry of the forget domain without compromising open-vocabulary generalization.

\subsection{Extended Implementation Details}

\textbf{Optimization and Training Regime.} Across all experiments, the model is optimized using AdamW for $10$ epochs with a learning rate of $1\times 10^{-5}$, a weight decay of $1\times 10^{-2}$, and a batch size of $64$. To ensure rigorous evaluation without state leakage, a fresh optimizer is instantiated for every sweep configuration, and every run initializes from the exact same pre-trained CLIP weights. We use the Nvidia A100 GPU with 40GB vRAM to conduct our experiments.

\textbf{TMS Hyperparameters.} For our geometric preference objective, we fix the temperature scaling at $\beta=0.1$, the retention margin at $m_{retain}=0$, and apply equal weighting to both the retention and confusion forces ($w_{retain} = w_{confuse} = 1$). The overall preference scaling weight ($\lambda_{pref}$) and the hard-mining confusion margin ($m_{confuse}$) are validated via grid search over $\lambda_{pref} \in \{0.5, 1.0, 2.0, 5.0\}$ and $m_{confuse} \in \{0.3, 0.5, 0.75\}$.

\textbf{Fisher Mask Computation.} The empirical Fisher Information Matrix is computed prior to the unlearning phase using $10$ minibatches drawn from the respective domain data loaders. When using a validation set and tuning the fisher threshold per domain individually, for multi-domain unlearning scenarios, we employ a rank-matched variant: mask sizes from per-domain validated thresholds are averaged, the sensitivity ratio $\rho$ is recomputed on the mixed forget set, and the top-$k$ parameters are selected to construct the final gating mask.

\subsection{Additional Baselines}

For additional baselines, we considered another recent unlearning method tested mostly on class unlearning. We adopt SEMU \cite{sendera2025semu} as a primary baseline, leveraging its SVD-based projection method for efficient, data-free machine unlearning. But since it relies mostly on having distinct subspaces, the domain entanglement in pre-trained VLMs make it extremely challenging for such methods to work off the shelf. Results on domainnet dataset, especially, were worse than baselines we consider in the main paper. On PACS, though inferior to the methods in the main paper, it did perform unlearning. Figure \ref{fig:domain_subspaces} clearly shows that the domains in the domainnet dataset are very entangled compared to the PACS dataset making SEMU approach fail.  

\begin{figure}[t] 
    \centering
    \resizebox{\columnwidth}{!}{\includegraphics[height=3.5cm, keepaspectratio]{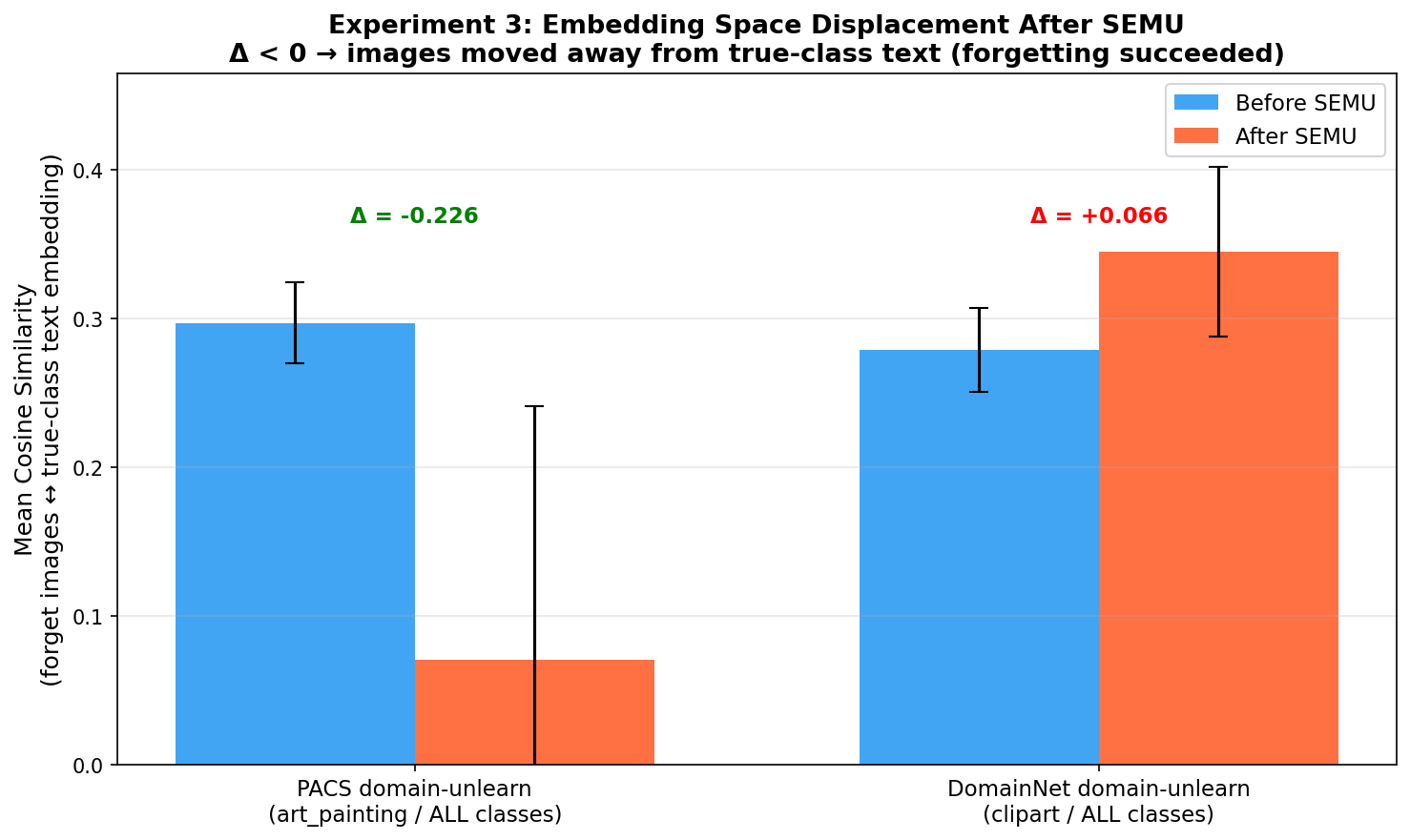}}
    
    \vspace{-12pt} 
    \caption{\small SEMU unlearning performance on PACS art\_painting domain and the Domainnet clipart domain.}
    \label{fig:SEMU}
    \vspace{-15pt} 
\end{figure}

\begin{figure}[t] 
    \centering
    \resizebox{\columnwidth}{!}{\includegraphics[height=3.5cm, keepaspectratio]{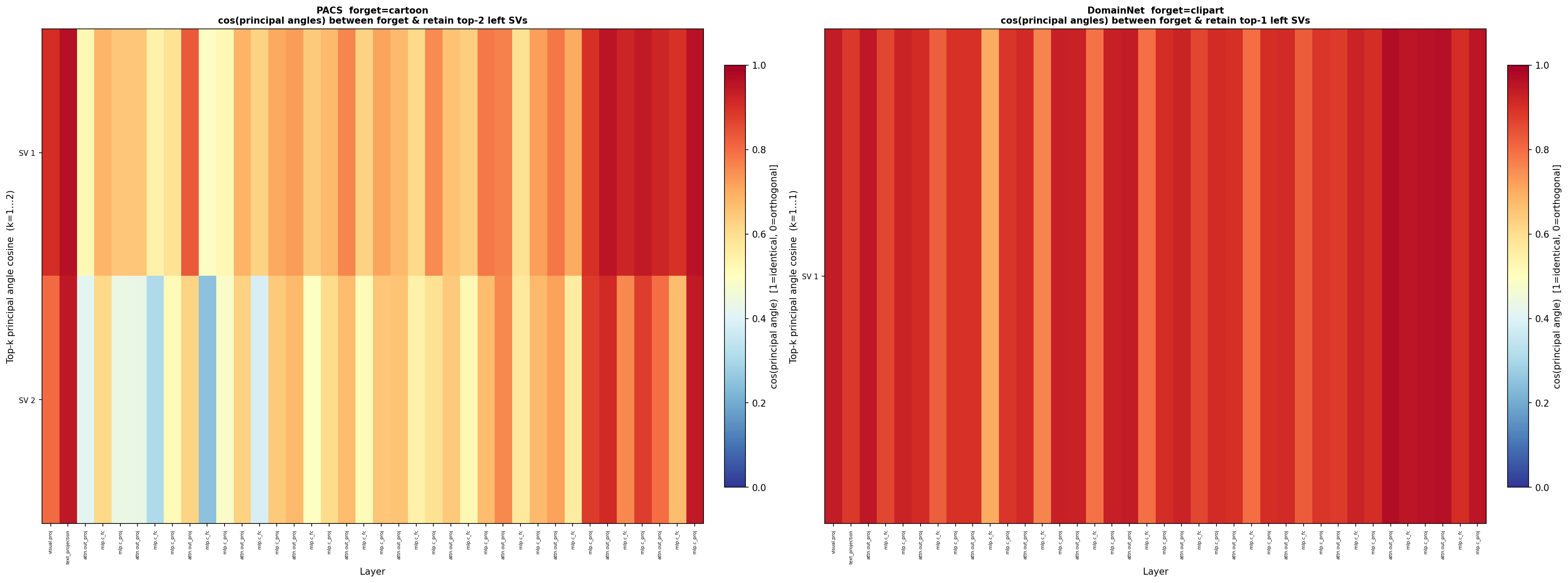}}
    
    \vspace{-12pt} 
    \caption{\small \textbf{Similarity between the top forget domain and top retain domain singular vectors.} The majority red amongst the domainnet domains implies that the orthogonality between domains do not exist making SEMU-based approaches very hard to get it working off the shelf.}
    \label{fig:heatmap}
    \vspace{-15pt} 
\end{figure}

\subsection{Additional Experiments}

To test the scalability of our method, we tested it on a stronger CLIP-based VLM with a ViT-L-14 backbone and compared it with the baseline and ADU. Tables \ref{tab:ViT-L-14}, \ref{tab:ViT-L-14-diffsetting}, and \ref{tab:TMS_ADU_ViTL} results clearly indicate that the proposed TMS objective improves forgetting by a large margin by conceding a very small margin in the retention, thereby achieving the best overall trade-off scores across the board.

\begin{table}[h!]
    \centering
    \caption{Comparison of Baseline and TMS Methods on a stronger pretrained ViT-L-14 backbone ($N=8$, $\gamma=0.5$).}
    \label{tab:ViT-L-14}
    \begin{tabular}{l | cc | cc | cc}
        \toprule
        \multirow{2}{*}{\textbf{Domain}} & \multicolumn{2}{c|}{\textbf{Forget Accuracy ($\downarrow$)}} & \multicolumn{2}{c|}{\textbf{Retain Accuracy ($\uparrow$)}} & \multicolumn{2}{c}{\textbf{Harmonic Mean ($\uparrow$)}} \\
        & Baseline & TMS & Baseline & TMS & Baseline & TMS \\
        \midrule
        Painting & 0.3016 & \textbf{0.2222} & \textbf{0.8603} & 0.8254 & 0.7710 & \textbf{0.8009} \\
        Real     & 0.8651 & \textbf{0.2937} & \textbf{0.8386} & 0.8069 & 0.2324 & \textbf{0.7533} \\
        Clipart  & 0.2714 & \textbf{0.1508} & \textbf{0.8354} & 0.8164 & 0.7784 & \textbf{0.8325} \\
        Sketch   & 0.1635 & \textbf{0.0841} & \textbf{0.8228} & 0.8074 & 0.8296 & \textbf{0.8582} \\
        \bottomrule
    \end{tabular}
\end{table}

\begin{table}[h!]
    \centering
    \caption{Comparison of Baseline and TMS on a stronger pretrained ViT-L-14 backbone($N=8$, $\gamma=1.0$).}
    \label{tab:ViT-L-14-diffsetting}
    \begin{tabular}{l | cc | cc | cc}
        \toprule
        \multirow{2}{*}{\textbf{Domain}} & \multicolumn{2}{c|}{\textbf{Forget Accuracy ($\downarrow$)}} & \multicolumn{2}{c|}{\textbf{Retain Accuracy ($\uparrow$)}} & \multicolumn{2}{c}{\textbf{Harmonic Mean ($\uparrow$)}} \\
        & Baseline & TMS & Baseline & TMS & Baseline & TMS \\
        \midrule
        Painting & 0.2222 & \textbf{0.1714} & \textbf{0.8566} & 0.8381 & 0.8153 & \textbf{0.8333} \\
        Real     & 0.1460 & \textbf{0.1206} & 0.8063 & \textbf{0.8095} & 0.8295 & \textbf{0.8430} \\
        Clipart  & 0.2079 & \textbf{0.1349} & \textbf{0.8360} & 0.8190 & 0.8134 & \textbf{0.8414} \\
        Sketch   & 0.1222 & \textbf{0.0746} & \textbf{0.8349} & 0.8212 & 0.8558 & \textbf{0.8702} \\
        \bottomrule
    \end{tabular}
\end{table}

\begin{table}[h!]
    \centering
    \caption{Comparison of Methods Across Domains ($N=8$, $\gamma=1.0$). Metrics are F: Forget Accuracy, R: Retain Accuracy, and HM: Harmonic Mean.}
    \label{tab:TMS_ADU_ViTL}
    \resizebox{\textwidth}{!}{
    \begin{tabular}{l | ccc | ccc | ccc | ccc}
        \toprule
        \multirow{2}{*}{\textbf{Method}} & \multicolumn{3}{c|}{\textbf{Painting}} & \multicolumn{3}{c|}{\textbf{Real}} & \multicolumn{3}{c|}{\textbf{Clipart}} & \multicolumn{3}{c}{\textbf{Sketch}} \\
        & \textbf{F ($\downarrow$)} & \textbf{R ($\uparrow$)} & \textbf{HM ($\uparrow$)} & \textbf{F ($\downarrow$)} & \textbf{R ($\uparrow$)} & \textbf{HM ($\uparrow$)} & \textbf{F ($\downarrow$)} & \textbf{R ($\uparrow$)} & \textbf{HM ($\uparrow$)} & \textbf{F ($\downarrow$)} & \textbf{R ($\uparrow$)} & \textbf{HM ($\uparrow$)} \\
        \midrule
        Baseline   & 0.2222 & 0.8566 & 0.8153 & 0.1460 & 0.8063 & 0.8295 & 0.2079 & 0.8360 & 0.8134 & 0.1222 & 0.8349 & 0.8558 \\
        ADU        & 0.2524 & \textbf{0.8815} & 0.8091 & 0.2683 & \textbf{0.8503} & 0.7866 & 0.2365 & \textbf{0.8624} & 0.8099 & 0.2365 & \textbf{0.8847} & 0.8196 \\
        TMS (Ours) & \textbf{0.1714} & 0.8381 & \textbf{0.8333} & \textbf{0.1206} & 0.8095 & \textbf{0.8430} & \textbf{0.1349} & 0.8190 & \textbf{0.8414} & \textbf{0.0746} & 0.8212 & \textbf{0.8702} \\
        \bottomrule
    \end{tabular}
    }
\end{table}

\subsection{Computational Efficiency}
Despite leveraging the empirical Fisher Information Matrix—a technique often associated with prohibitive computational overhead—our framework is designed for rapid deployment. Rather than continuously re-evaluating parameter sensitivity, the Fisher mask is computed strictly once prior to the unlearning phase using a minimal subset of data ($10$ minibatches). Furthermore, the structural tension induced by our Targeted Manifold Scattering (TMS) objective extracts maximal unlearning signal from every individual image via dynamic hard-mining. This geometric efficiency drives exceptionally fast convergence, allowing the entire domain erasure process to complete in just $10$ epochs on a highly constrained few-shot dataset ($K=8$). Combined with fully frozen, pre-computed text features and sparse updates restricted to the visual encoder, our approach achieves open-vocabulary domain unlearning at a fraction of the computational cost required by standard retraining or continuous prompt-tuning baselines.

\newpage

\end{document}